\documentclass[letterpaper, 10 pt, conference]{ieeeconf}  

\IEEEoverridecommandlockouts                              
\usepackage{cite}
\usepackage{amsmath,amssymb,amsfonts}
\usepackage{algorithmic}
\usepackage{algorithm}
\usepackage{graphicx}
\usepackage{textcomp}
\usepackage{xcolor}
\usepackage[hidelinks]{hyperref}

\usepackage{booktabs}
\usepackage{adjustbox}
\usepackage{multirow}
\usepackage{tikz}
\usetikzlibrary{arrows.meta, positioning, bending}

\begin{document}

\title{\LARGE \bf Synthetic Data Generation and Vision-based Wrinkle and Keypoint Detection for Bimanual Cloth Manipulation}

\author{Ariel Herrera$^{1}$, Xueyang Kang$^{2}$, and Atal Anil Kumar$^{3}$
\thanks{$^{1}$Ariel Herrera is with the Department of Engineering, University of Luxembourg, Esch-sur-Alzette, Luxembourg.
        {\tt\small arielmh.herrera@gmail.com}}%
\thanks{$^{2}$Xueyang Kang is with the School of Electrical and Electronic Engineering, Nanyang Technological University, Singapore.
        {\tt\small xueyang.kang@ntu.edu.sg}}%
\thanks{$^{3}$Atal Anil Kumar is with the Université de Lorraine, Arts et Metiers Institute of Technology, LCFC, F-57070 Metz, France
        {\tt\small atal-anil.kumar@univ-lorraine.fr}}%
}

\maketitle

\begin{abstract}
Robotic manipulation of textiles remains challenging because continuous deformation and self-occlusions hinder the robust visual perception required to estimate the cloth's state. To address the lack of annotated real-world data, we developed a Blender-based synthetic pipeline exporting auto-annotated keypoints, and combined manually labeled renders with real-world data to train a wrinkle detector. We present a perception framework integrating a CNN for permutation-invariant keypoint detection and a YOLOv8-OpenCV pipeline to extract grasping points from structural wrinkles. A proposed bimanual algorithm uses this system to stretch fully folded garments via wrinkles, transitioning to keypoint-based ironing once corners emerge. The keypoint model achieves a Mean Position Error (MPE) of 1.7615 $\pm$ 0.7461 pixels. The perception system transfers to physical fabrics without fine-tuning, outperforming baselines that fail in high-occlusion states or yield false positives on severe folds. The code and datasets are available at: \url{https://github.com/arielherreraaguiar/Grasping-Points-Detection}
\end{abstract}

\begin{keywords}
Synthetic Data, Deformable Object Manipulation, Visual Perception, Keypoint Detection, Wrinkle Detection, Bimanual Manipulation, Sim-to-Real.
\end{keywords}

\section{Introduction}
Robotic manipulation of textiles is challenging because infinite degrees of freedom and continuous deformation create severe self-occlusions, rendering standard pose estimation ineffective. Successful bimanual unfolding relies on visual perception to reliably extract structural features like corners (keypoints) and internal folds (wrinkles).

Training these perception models requires massive datasets. Because manual labeling is labor-intensive, researchers increasingly use domain-randomized physics simulations to generate synthetic datasets, enabling Sim-to-Real transfer without manual data collection.

Despite these advances, existing frameworks struggle with completely folded garments. Methods relying exclusively on keypoint detection fail when external corners are entirely occluded, leaving the robot without valid grasping points to initiate manipulation.

This paper addresses this gap with a vision-driven bimanual framework targeting fully folded states. The system isolates structural wrinkles to compute precise grasping points for initial stretching, transitioning to point-based tracking once collinear keypoints are exposed.

The primary contributions of this work are:
\begin{itemize}
    \item A domain-randomized Blender pipeline simulating cloth physics and exporting auto-annotated keypoints, supplemented with real-world data for wrinkle detection.
    \item A real-time perception framework integrating a CNN for keypoint detection (achieving an MPE of 1.7615 pixels) and a YOLOv8-OpenCV pipeline extracting grasping points from wrinkles.
    \item A bimanual manipulation algorithm transitioning from wrinkle-based stretching to keypoint-based ironing, establishing a baseline strategy for future Reinforcement Learning.
\end{itemize}

\section{Related Work}
Visual perception for deformable object manipulation has seen significant advancements, with recent works focusing on learning robust representations for cloth understanding and manipulation. Lips et al.~\cite{lips2022} proposed a Blender-based Sim-to-Real framework for deformable object pose estimation using keypoints; however, their dataset is limited primarily to flat cloth configurations and focuses mainly on folding-oriented keypoint localization without addressing highly occluded or non-flat unfolding scenarios. Similarly, Li et al.~\cite{li2023yolo} utilized YOLOv5-based object detection for wrinkle and corner feature detection in fabrics, demonstrating strong performance for visible corner localization but exhibiting limitations when cloth structures become highly folded or self-occluded. Earlier geometric approaches have also explored grasp point reasoning for cloth manipulation. Martínez and Ruiz-del-Solar~\cite{martinez2013grasp} extracted garment contours and employed polygonal approximation with curvature analysis to detect graspable regions such as corners and edges. While effective for visible boundary structures, contour-driven methods often struggle when key grasping regions are hidden within complex internal folds. 

More recently, cloud-assisted robotic systems such as Dressing-as-a-Service~\cite{xu2024dressing} introduced scalable bimanual assistive manipulation frameworks that rely heavily on accurate garment state estimation and perception robustness. Beyond deformable manipulation, recent advances in semantic scene understanding and 3D perception provide complementary insights for robust cloth representation learning. Robust semantic association strategies for dynamic environments have been explored in semantic SLAM systems~\cite{kang2019robust}, demonstrating the importance of stable feature association under partial occlusions and structural ambiguities. Interactive semantic embedding methods for 3D segmentation~\cite{kang2026few} further highlight the effectiveness of semantic-aware feature representations for sparse and ambiguous visual structures, which are highly relevant for identifying occluded garment regions. In addition, hierarchical point-patch fusion frameworks with adaptive codebook representations~\cite{kang2026hierarchical} have shown strong capability in capturing fine-grained geometric anomalies and local structural variations, offering useful insights for modeling complex wrinkle distributions and folded cloth geometries. Recent generative scene understanding approaches based on panorama and video diffusion~\cite{kang2025look} also demonstrate the potential of multi-view contextual reasoning to recover structurally consistent representations from incomplete observations. Motivated by these limitations and recent advances, our approach jointly integrates keypoint reasoning with internal wrinkle-aware perception, enabling robust bimanual garment unfolding even when corners are severely occluded or hidden within complex folded configurations. For more deformable manipulation simulations, please refer to the survey paper \cite{wong2025survey}.

\section{Method}

\subsection{Synthetic Data Generation}
To train the neural networks without manual annotation, we implemented an automated pipeline using Blender \cite{blender}. We generated a large-scale dataset to ensure robust Domain Randomization. 

The cloth simulation in Blender is governed by a mass-spring physics model based on Newton's second law. In our pipeline, the garment is dropped from a random height. Upon impact with the table collision plane, the solver computes the reaction forces, causing the mesh to buckle and generate highly randomized, realistic deformations and folds.

\begin{figure}[htbp]
\centerline{\includegraphics[width=0.8\linewidth]{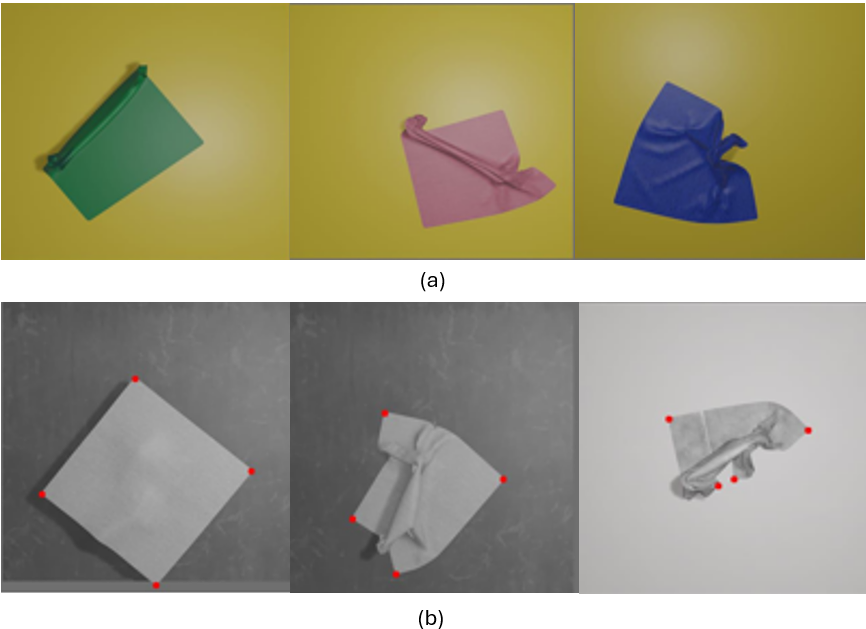}}
\caption{Examples from the synthetic dataset generated in Blender. (a) Top row: RGB renders showcasing domain randomization, including varying textures, lighting, and complex fold topologies. (b) Bottom row: Corresponding automated ground-truth annotations of the garment's corner keypoints.}
\label{fig:dataset_samples}
\end{figure}

The script was extended to run 500 episodes, resulting in 5000 labeled images. During each episode, we randomized: (i) cloth textures and optional color tints, (ii) table textures, (iii) initial drop positions and orientations, and (iv) lighting and camera poses. We utilized 20 distinct cloth textures (cycled with color tints) and 8 high-resolution table textures from public libraries \cite{ambientcg, cgbookcase, polyhaven}. Exemplar samples of the generated dataset are shown in Fig. \ref{fig:dataset_samples}. The data generation process is detailed in Algorithm \ref{alg:data_gen}.

\begin{algorithm}
\caption{Automated Dataset Generation Pipeline}
\label{alg:data_gen}
\begin{algorithmic}[1]
\REQUIRE $N_{episodes} \leftarrow 500$, Environment Parameters (Textures, Lighting, Drops)
\ENSURE Dataset $D$ containing pairs $(I, Y)$
\STATE $D \leftarrow \emptyset$
\FOR{$i = 1$ to $N_{episodes}$}
    \STATE $S_{init} \leftarrow \text{InitializeSimulationEnvironment}()$
    \STATE $S_{random} \leftarrow \text{ApplyDomainRandomization}(S_{init},$
    \STATE \quad $\text{Textures}, \text{Lighting}, \text{Drops})$
    \STATE $M_{sim} \leftarrow \text{RunPhysicsSimulation}(S_{random})$
    \STATE $I_{img} \leftarrow \text{RenderImage}(M_{sim})$
    \STATE $Y_{labels} \leftarrow \text{ExtractKeypointsAndSegmentation}(M_{sim})$
    \STATE $D \leftarrow D \cup \{(I_{img}, Y_{labels})\}$
\ENDFOR
\RETURN $D$
\end{algorithmic}
\end{algorithm}

\subsection{Keypoints Detection}
The keypoint detection model utilizes a custom Convolutional Neural Network (CNN) architecture trained to output a single-channel spatial heatmap. Unlike standard regression models that force a specific order of corners, our approach is permutation-invariant; each visible corner of the garment is naturally represented as a Gaussian peak within the generated heatmap.

During inference, the network extracts multiple peak candidates from the heatmap. To achieve sub-pixel accuracy for the 2D coordinates, we apply a local soft-argmax (centroid) refinement over a predefined window around each peak. The refined coordinates are then mapped back to the original high-resolution image space. 

Because the network can sometimes produce noisy or overlapping heatmaps, multiple peak candidates may cluster around a single physical corner. To resolve this, we implemented a greedy distance-based Non-Maximum Suppression (NMS) filter in the image space. If two predicted keypoints are closer than a defined pixel radius, the algorithm evaluates their confidence scores and suppresses the one with the lower score. This ensures that only highly confident, distinct keypoints are preserved, dynamically capping the output to a maximum of four corners.

\subsection{Wrinkle Detection}
When the cloth is highly folded, keypoints are often completely occluded. To train the wrinkle detection model, we combined custom synthetic Blender data with the Wrinkle Detector 2.0 Dataset from Roboflow \cite{roboflow_wrinkle}. From the total dataset of 5000 synthetic images generated in Blender, a subset of 500 images was annotated using the AnyLabeling tool \cite{anylabeling} to ensure accurate bounding box representations of folds before being merged. The detection model utilizes a YOLOv8 architecture \cite{yolov8} trained to predict bounding boxes around potential folds in the fabric.

To extract grasping points from the bounding boxes, we process the detections using OpenCV. For each Region of Interest (ROI), the image is converted to grayscale, and edges are extracted using the Canny detector. Morphological closing and dilation are then applied to connect fragmented edges, followed by contour extraction. To isolate the main fold, an ellipse is fitted to each contour to measure its major axis. The contour with the longest major axis across all ROIs is identified as the primary wrinkle, and a binary mask is generated.

Finally, to find the 2D grasping coordinates, we compute the convex hull of the wrinkle mask. By calculating the Euclidean distances between all points on the hull, the algorithm finds the two points separated by the maximum distance. These endpoints define the longest line across the wrinkle and serve as the target grasping points for the robotic arms.

\subsection{Fused Bimanual Manipulation Algorithm}
\begin{figure}[htbp]
\centerline{\includegraphics[width=\linewidth]{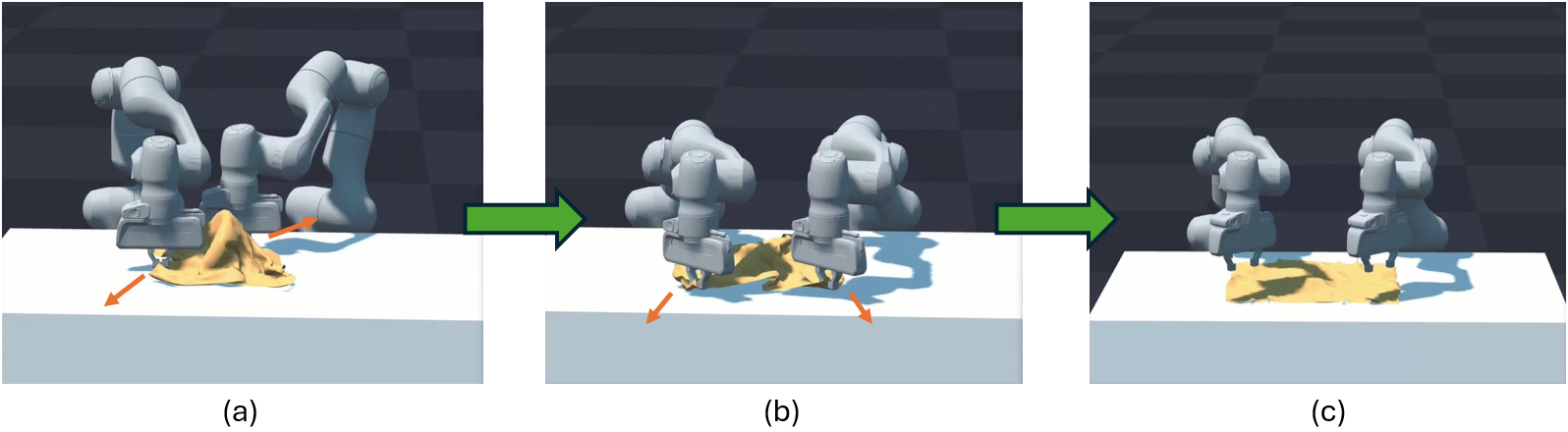}}
\caption{Conceptual bimanual manipulation sequence simulated using the Newton physics engine. Orange arrows indicate the planned trajectories for the robot end-effectors. (a) Wrinkle-based grasping and stretching for a completely folded garment. (b) Transition to keypoint-based ironing once collinear corners emerge. (c) The final fully unfolded state.}
\label{fig:concept_sim}
\end{figure}

To integrate the visual modalities, we propose a state-machine algorithm (Algorithm \ref{alg:fusion}) to govern the unfolding process. Fig. \ref{fig:concept_sim} illustrates this sequence in a conceptual simulation rendered using the Newton physics engine \cite{newton_physics}. 

The routine takes a continuous camera feed of a completely folded cloth as input. Initially, when no keypoints are visible, the system extracts grasping points from the structural wrinkle mask to stretch the fabric outward (Fig. \ref{fig:concept_sim}a). This action repeats until at least two collinear keypoints emerge. The robots then transition to a keypoint-guided ironing motion (Fig. \ref{fig:concept_sim}b). The loop continues until all four keypoints are localized, resulting in a fully unfolded garment (Fig. \ref{fig:concept_sim}c).

\begin{algorithm}
\caption{Fused Bimanual Manipulation for Cloth Unfolding}
\label{alg:fusion}
\begin{algorithmic}[1]
\REQUIRE Continuous camera feed for a completely folded cloth
\ENSURE Cloth fully unfolded (4 keypoints detected)
\WHILE{Detected Keypoints $< 4$}
    \STATE Capture image $I$
    \STATE Process $I$ to detect Keypoints $P$ and Wrinkle mask $M_w$
    \IF{$|P| < 2$ \textbf{or not} collinear($P$)}
        \STATE \textit{// Wrinkle Stretching Action}
        \STATE Extract longest wrinkle line $L_w$ from $M_w$
        \STATE Compute midpoint $m_w$ of $L_w$
        \STATE Compute perpendicular bisector $B_w$ at $m_w$
        \STATE Find intersections $P_1, P_2$ of $B_w$ with cloth edges
        \STATE Robot 1 grasps $P_1$, Robot 2 grasps $P_2$
        \STATE Move arms outwards to stretch cloth
    \ELSE
        \STATE \textit{// Ironing Action}
        \STATE \textbf{assert} $|P| \ge 2$ \textbf{and} collinear($P$)
        \STATE Robot 1 and 2 grasp the detected keypoints in $P$
        \STATE Move end-effectors forward/backward to smooth cloth
    \ENDIF
\ENDWHILE
\end{algorithmic}
\end{algorithm}

\section{Experiments and Results}
We evaluated the keypoint detection model by calculating the Mean Position Error (MPE) between the predicted 2D pixel coordinates and the ground truth from our synthetic test set. 

\begin{figure}[htbp]
\centerline{\includegraphics[width=\linewidth]{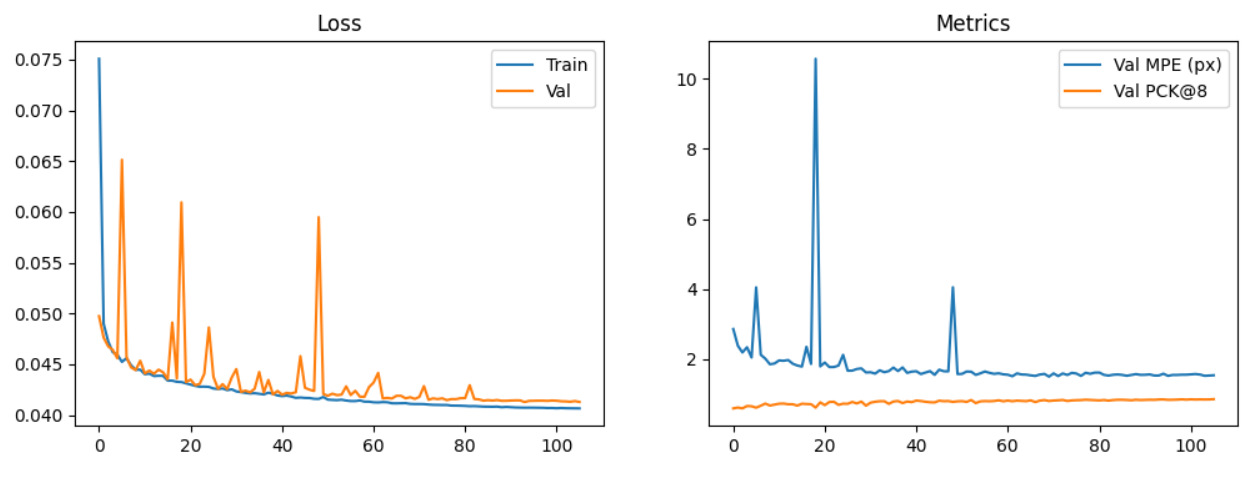}}
\caption{Training and validation metrics over epochs. The left plot demonstrates the successful convergence of the training and validation loss, while the right plot illustrates the rapid stabilization of the validation Mean Position Error (MPE) and PCK@8 metrics.}
\label{fig:training_metrics}
\end{figure}

Fig. \ref{fig:training_metrics} shows the loss and evaluation metrics across all training epochs. Both training and validation loss converged steadily, showing no signs of overfitting. Furthermore, the validation MPE and PCK@8 (Percentage of Correct Keypoints) stabilized quickly, confirming an effective training process.

\begin{figure}[htbp]
\centerline{\includegraphics[width=\linewidth]{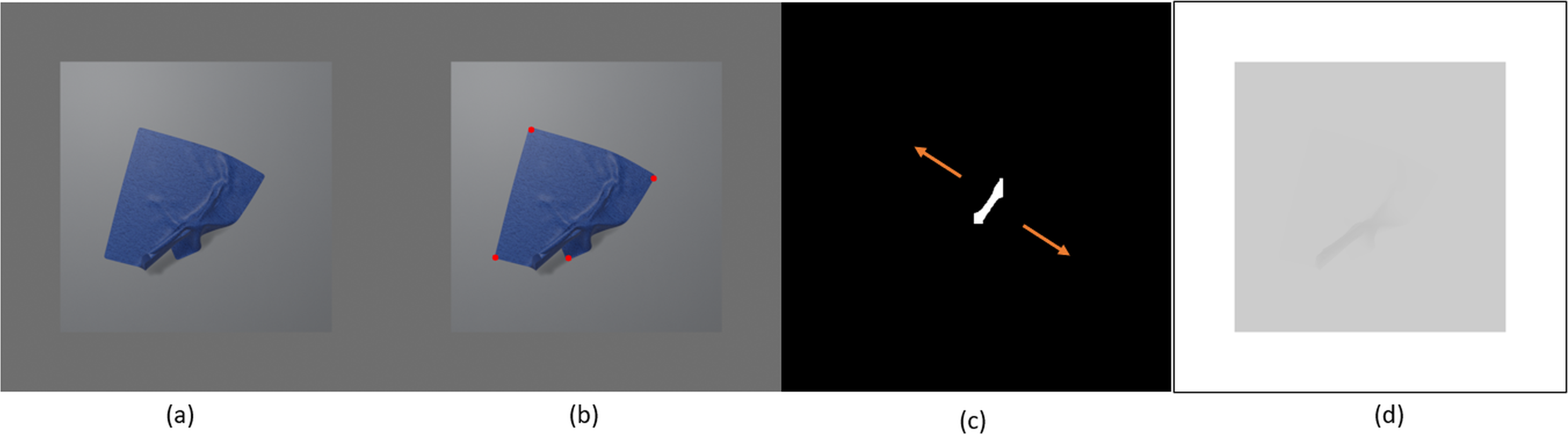}}
\caption{Vision framework evaluation on synthetic data. (a) Synthetic cloth generated in Blender. (b) Keypoint detection output. (c) Wrinkle segmentation with orange arrows indicating the directional vectors the bimanual robots must follow to unfold the cloth. (d) Corresponding depth map of the synthetic garment.}
\label{fig:results_sim}
\end{figure}

The model achieves a final MPE of 1.7615 $\pm$ 0.7461 pixels. This low error and variance confirm that the network accurately localizes features despite complex cloth deformations. Fig. \ref{fig:results_sim} visualizes the keypoint extraction and the wrinkle-based grasping point calculation on the synthetic test set.

\begin{figure}[htbp]
\centerline{\includegraphics[width=\linewidth]{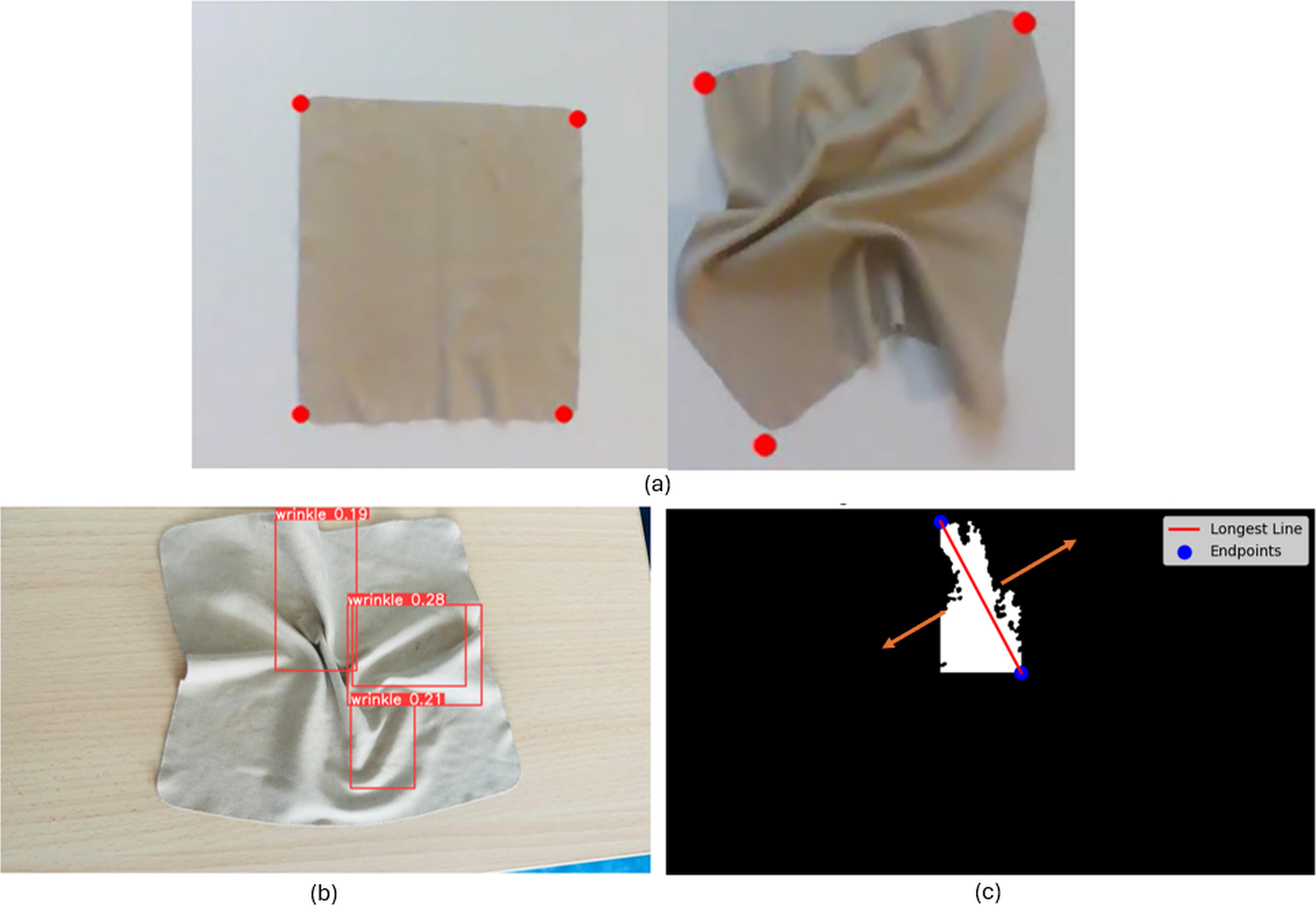}}
\caption{Sim-to-Real transfer performance on physical fabrics. (a) Keypoint detection on real cloth in both fully unfolded and folded states. (b) Raw wrinkle detection bounding boxes. (c) Post-processing that isolates the longest structural wrinkle, performs binary segmentation, extracts the maximal fold line via convex hull (blue line), and generates the grasping points with unfolding directions (orange arrows).}
\label{fig:results_real}
\end{figure}

Our models transfer to the real world without any fine-tuning. The system robustly detects keypoints and wrinkles on physical fabrics, validating the Sim-to-Real pipeline. Fig. \ref{fig:results_real} demonstrates this transfer, showing the framework predicting accurate unfolding trajectories on real textiles.

\subsection{Baseline Comparisons}
To validate the robustness of our approach, we compared our results against state-of-the-art baselines. Table \ref{tab:baselines} summarizes the Mean Position Error (MPE) for each method, particularly evaluating performance in completely folded scenarios where keypoints are occluded.

\begin{table}[htbp]
\caption{Mean Position Error (MPE) Comparison}
\begin{center}
\resizebox{\columnwidth}{!}{%
\begin{tabular}{|l|c|p{4cm}|}
\hline
\textbf{Method} & \textbf{MPE (Pixels)} & \textbf{Primary Limitations in Folded States} \\
\hline
Lips et al. \cite{lips2022} & 26.0830 $\pm$ 27.1104 & Unstable behavior with severe folds; lacks wrinkle detection capabilities. \\
\hline
Cheng Li \cite{li2023yolo} & 10.4520 $\pm$ 4.1205 & Generates false positives (YOLOv5 misclassifies complex folds as keypoints). \\
\hline
\textbf{Ours} & \textbf{1.7615 $\pm$ 0.7461} & \textbf{Robust grasping point extraction via longest wrinkle isolation.} \\
\hline
\end{tabular}%
}
\label{tab:baselines}
\end{center}
\end{table}

\begin{figure}[htbp]
\centerline{\includegraphics[width=0.8\linewidth]{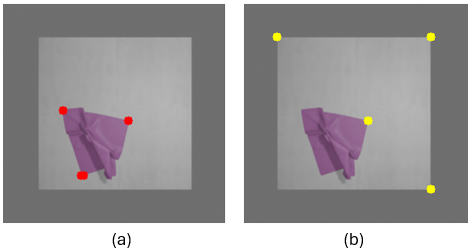}}
\caption{Qualitative comparison of keypoint detection on folded cloth. (a) Accurate keypoint localization using our proposed model. (b) Incorrect and unstable keypoint detection when using the baseline model proposed by Lips et al. \cite{lips2022}.}
\label{fig:baseline_comparison}
\end{figure}

Fig. \ref{fig:baseline_comparison} illustrates that while baselines succeed on partially unfolded fabrics, they fail in completely folded states. Lips et al. lack a wrinkle-based fallback, resulting in high keypoint prediction errors on severe folds. Similarly, Cheng Li's YOLOv5 approach generates false positive keypoints when only wrinkles are visible. In contrast, our method isolates structural wrinkles to extract reliable grasping points, maintaining accuracy even when keypoints are fully occluded.

\section{Conclusion and Future Work}
This paper presented a vision-driven perception framework for bimanual cloth manipulation. By generating a domain-randomized synthetic dataset in Blender and supplementing it with real-world data for wrinkle detection, we developed models capable of extracting garment keypoints and structural wrinkles. The keypoint detection CNN achieved an MPE of 1.7615 pixels on the synthetic test set, and the complete perception pipeline transfers to physical fabrics without fine-tuning. Compared to baseline methods, our approach accurately identifies grasping points even in completely folded states. Additionally, we proposed a manipulation algorithm that transitions from wrinkle-based stretching to keypoint-based ironing.

For future work, we will integrate this perception framework with Reinforcement Learning (RL) in the Newton physics engine to train bimanual unfolding policies in simulation. Subsequently, we plan to transfer these learned policies to a physical robotic setup using Sim-to-Real techniques to evaluate real-world grasp quality and unfolding success rates.

\section*{ACKNOWLEDGMENT}
The authors would like to thank the ORION program for its contribution to the funding of A.H.'s research internship, carried out at the École Nationale d'Ingénieurs de Metz (ENIM), University of Lorraine. This work has also benefited from a government grant managed by the Agence Nationale de la Recherche with the reference ANR-20-SFRI-0009.

\bibliographystyle{IEEEtran}
\bibliography{references}

\end{document}